\documentclass{preprint}

\usepackage{lipsum} 
\usepackage{xcolor}
\usepackage{booktabs}
\usepackage{graphicx} 
\usepackage[numbers,comma,sort&compress,super]{natbib}
\usepackage[colorlinks=true]{hyperref}
\usepackage{makecell}
\usepackage{tabularx}

\title{Closing the gap between open-source and commercial large language models for medical evidence summarization}
\author[1]{Gongbo Zhang}
\author[2]{Qiao Jin}
\author[3]{Yiliang Zhou}
\author[4]{Song Wang}
\author[1]{Betina R. Idnay}
\author[5]{Yiming Luo}
\author[5]{Elizabeth Park}
\author[5]{Jordan G. Nestor}
\author[6]{Matthew E. Spotnitz}
\author[7,8,9]{Ali Soroush}
\author[3,10]{Thomas Campion}
\author[2]{Zhiyong Lu}
\author[1,*]{Chunhua Weng}
\author[3,10,*]{Yifan Peng}
\affil[1]{Department of Biomedical Informatics, Columbia University, New York, NY, USA}
\affil[2]{National Center for Biotechnology Information, National Library of Medicine, National Institutes of Health, Bethesda, MD, USA}
\affil[3]{Department of Population Health Sciences, Weill Cornell Medicine, New York, NY, USA}
\affil[4]{Cockrell School of Engineering, The University of Texas at Austin, Austin, TX, USA}
\affil[5]{Department of Medicine, Columbia University, New York, NY, USA}
\affil[6]{Office of the Director, National Institutes of Health, Bethesda, MD, USA}
\affil[7]{Division of Data-Driven and Digital Medicine, Icahn School of Medicine at Mount Sinai, New York, NY, USA}
\affil[8]{Charles Bronfman Institute for Personalized Medicine, Icahn School of Medicine at Mount Sinai, New York, NY, USA}
\affil[9]{Henry D. Janowitz Division of Gastroenterology, Icahn School of Medicine at Mount Sinai, New York, NY, USA}
\affil[10]{Clinical \& Translational Science Center, Weill Cornell Medicine, New York, NY, USA}
\affil[*]{Corresponding: Yifan Peng, \url{yip4002@med.cornell.edu}; Chunhua Weng, \url{cw2384@cumc.columbia.edu}}

\begin{document}

\maketitle

\begin{abstract}
Large language models (LLMs) hold great promise in summarizing medical evidence. Most recent studies focus on the application of proprietary LLMs. Using proprietary LLMs introduces multiple risk factors, including a lack of transparency and vendor dependency. While open-source LLMs allow better transparency and customization, their performance falls short compared to proprietary ones. In this study, we investigated to what extent fine-tuning open-source LLMs can further improve their performance in summarizing medical evidence. Utilizing a benchmark dataset, MedReview, consisting of 8,161 pairs of systematic reviews and summaries, we fine-tuned three broadly-used, open-sourced LLMs, namely PRIMERA, LongT5, and Llama-2. Overall, the fine-tuned LLMs obtained an increase of 9.89 in ROUGE-L (95\% confidence interval: 8.94-10.81), 13.21 in METEOR score (95\% confidence interval: 12.05-14.37), and 15.82 in CHRF score (95\% confidence interval: 13.89-16.44). The performance of fine-tuned LongT5 is close to GPT-3.5 with zero-shot settings. Furthermore, smaller fine-tuned models sometimes even demonstrated superior performance compared to larger zero-shot models. The above trends of improvement were also manifested in both human and GPT4-simulated evaluations. Our results can be applied to guide model selection for tasks demanding particular domain knowledge, such as medical evidence summarization.
\end{abstract}


\section{Introduction}\label{introduction}

Medical evidence plays a critical role in healthcare decision-making. Particularly, systematic reviews and meta-analyses of randomized controlled trials (RCTs) are considered the gold standard for generating robust medical evidence.\cite{Peng2023-zh, Concato2000-np} However, systematically reviewing multiple RCTs is laborious and time-consuming.\cite{Borah2017-ah} It requires retrieving relevant studies, appraising the evidence quality, and synthesizing findings. Meanwhile, systematic reviews frequently become obsolete upon publication, primarily due to protracted review processes. The delay is exacerbated by the exponential increase in scientific discoveries, exemplified by over 133,000 new clinical trials registered at ClinicalTrials.gov since 2020.\cite{NLM2024-qw} As such, it is imperative to establish an efficient, reliable, and scalable automated system to streamline and accelerate systematic reviews.

Typically, systematic reviews include quantitative and qualitative reports,\cite{Page2021-nl} the former being a statistical meta-analysis of relevant clinical trials, and the latter being a concise narrative explanation of the quantitative results.\cite{Wallace2021-zh, Tang2023-oa} Language generation technologies could potentially be employed to auto-generate such narratives, yet they have not been widely applied for medical evidence summarization.\cite{Tang2023-oa} Text summarization has attracted research attention for decades. Earlier techniques relied on extracting key phrases and sentences through rules or statistical heuristics like word frequency and sentence placement.\cite{Barzilay2002-nv, Pivovarov2015-km, Zweigenbaum2007-jj, Li2010-rc, Demner-Fushman2007-tk} These methods, however, struggled with comprehending context and generating cohesive summaries. A significant shift occurred with the adoption of neural network-based methods, enhanced by attention mechanisms.\cite{Gu2021-ni, Guo2021-qv, Xiao2021-et, Zhang2020-fq, Lewis2019-qk, Devlin2018-ct, Mrabet2020-xj} These mechanisms enable the model to concentrate on various input segments and understand long-range connections between text elements. This advancement allows for a deeper grasp of context, leading to smoother and more precise summaries.

Recent advancements in generative Artificial Intelligence (AI), notably large language models (LLMs), have shown tremendous potential in comprehending and generating natural language.\cite{Peng2023-zh, Singhal2023-jc} While these generalist models perform well across diverse tasks, they fail to capture in-depth domain-specific knowledge, particularly in biomedicine.\cite{Zack2023-jd, Jin2023-pk} Furthermore, despite the relevant superior performances compared to open-source alternatives,\cite{Tang2023-oa, Jiang2024-jw, Ouyang2022-ww, Touvron2023-we, OpenAI2023-cn} the disadvantages of closed-source models were rarely discussed or even mentioned.\cite{Nosek2015-yy, Zhang2024-bu} The lack of transparency of closed-source models makes it challenging to understand the model behavior and troubleshoot customized variants. Moreover, reliance on closed-source models raises the risks associated with changes in service terms or discontinuation of services, which pose a critical threat to long-term projects. Open-source models provide a promising solution to mitigate the above risks.

While multiple open-source model architectures were proposed for either general-purpose foundation models or specifically for text summarization, few have been optimized to synthesize medical evidence. Gutierrez et al. compared the in-context learning performance of GPT-3 with fine-tuned BERT models on Named-Entity Recognition tasks in the biomedical domain.\cite{Gutierrez2022-xu} Tang et al. assessed zero-shot GPT-3.5 and ChatGPT on summarizing Cochrane review abstracts.\cite{Tang2023-oa} The above related studies, however, did not focus on fine-tuning open-source models for medical evidence summarization. It's also unclear to what degree the optimization strategies, e.g., few-shot learning or fine-tuning, can help bridge the performance gap between open-source models and cutting-edge closed-source alternatives. To quantitively assess fine-tuning technologies to enhance open-source LLMs for medical evidence summarization, we experimented with three broadly used open-sourced LLMs: PRIMERA,\cite{Xiao2021-et} LongT5,\cite{Guo2021-qv} and Llama-2,\cite{Touvron2023-we} including both architectures designed specifically for text summarization and architectures of generalist foundation models. Fine-tuning these models is challenging due to the substantial requirement for computational resources and the risk of catastrophic forgetting, a phenomenon in machine learning where the performance degrades on tasks where the LLM initially performed well.\cite{Tadros2022-ul} To address these issues, we employed Low-Rank Adaptation (LoRA),\cite{Hu2021-ey} which is a parameter-efficient fine-tuning method focusing on updating only a minimal amount of model parameters during the fine-tuning process.

\begin{figure}
    \centering
    \includegraphics[width=.8\linewidth]{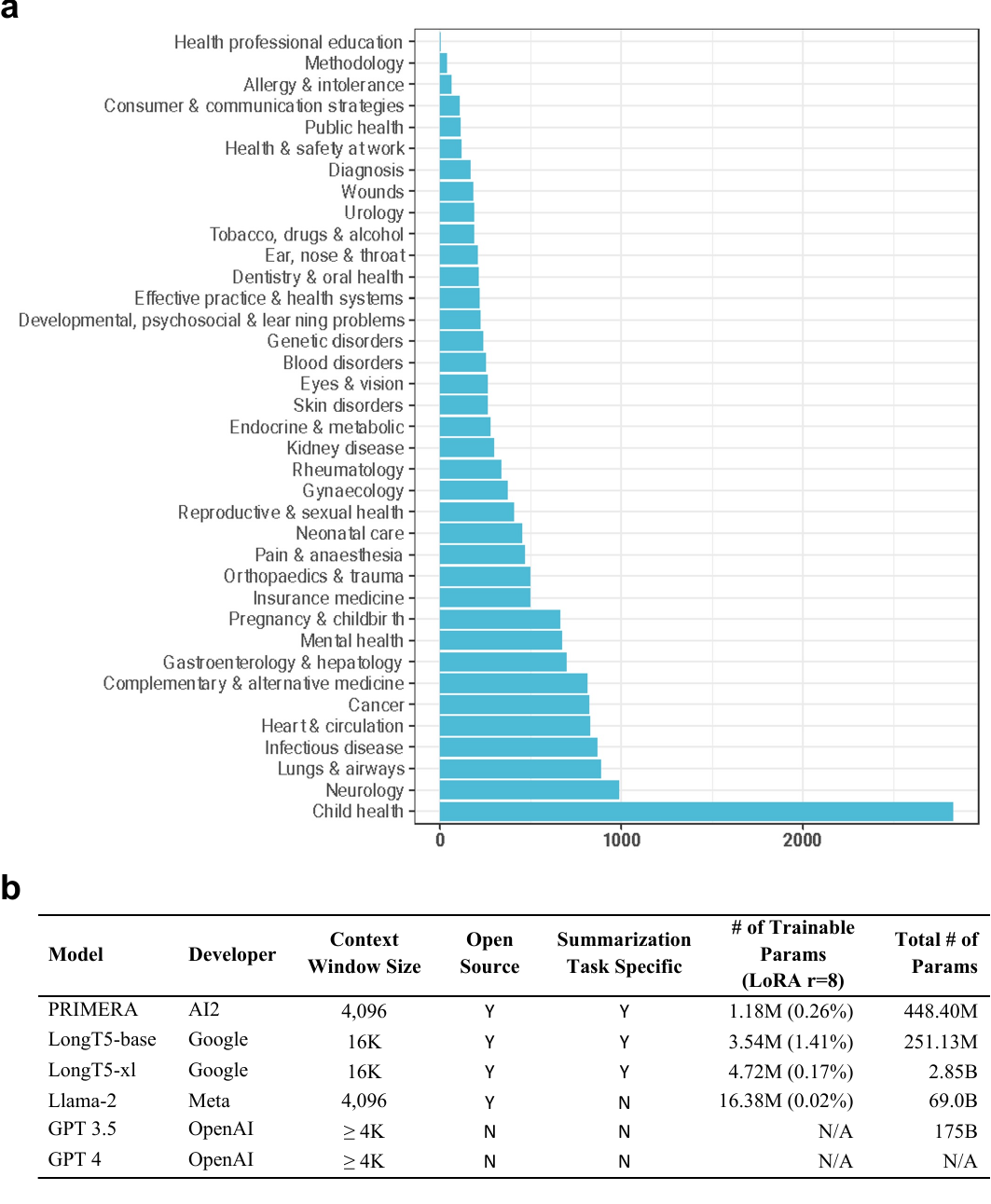}
    \caption{Overview of topic distribution of the MedReview dataset and LLMs in this study. \textbf{a,} Topic distribution of the MedReview dataset. \textbf{b,} Choice of LLMs in this study.}
    \label{fig:overview}
\end{figure}

To facilitate future studies on leveraging LLMs for medical evidence summarization, we present a benchmark dataset, MedReview, consisting of 8,161 pairs of meta-analysis results and narrative summaries from the Cochrane Library,\cite{Cochrane-wg} published on 37 topics between April 1996 and June 2023. The dataset consists of training, validation, and test sets (see details in the Method section and Supplement Table \ref{tab:dataset}). This collection is an extension of our previous study of evaluating LLMs for evidence summarization\cite{Tang2023-oa} and covers a wider range of specialties and writing styles, which highlight common text summarization challenges (Figure \ref{fig:overview}a).

\section{Results}\label{results}

\subsection{Comparison of different LLMs in automatic evaluations}\label{comparison-of-different-llms-in-automatic-evaluations}

\begin{figure}
    \centering
    \includegraphics[width=\linewidth]{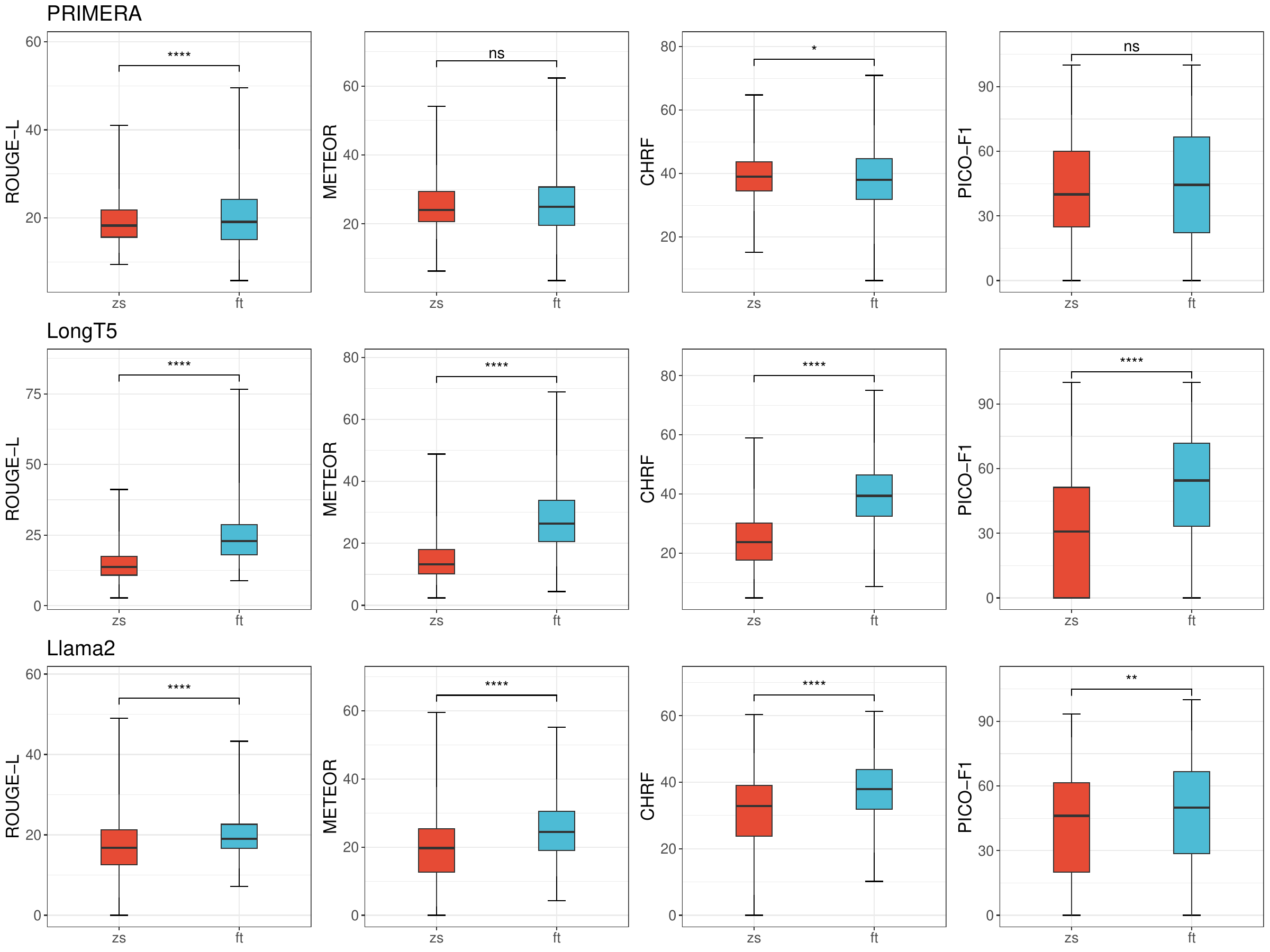}
    \caption{Performance of different medical evidence summarization systems in automatic evaluations. The p-value was calculated using a paired t-test to determine the statistical significance of the difference between the models. FT - fine-tuning; ZS - zero-shot learning; * - $p<0.05$; ** - $p<0.01$; *** - $p<0.001$; **** - $p<0.0001$; ns - Not significant.}
    \label{fig:automatic}
\end{figure}

First, we fine-tuned PRIMERA, LongT5, and Llama-2 using the LoRA method (Figure~\ref{fig:overview}b). We observed that fine-tuning considerably improved the performance of most models ($p < 0.01$, Figure~\ref{fig:automatic}). Specifically, LongT5 models benefited the most from fine-tuning, which led to an increase from 14.72 to 24.61, 15.06 to 28.27 in METEOR, 15.15 to 38.81 in CHRF, and 36.05 to 51.43 in PICO-F1 (Supplementary Table~\ref{tab:automatic}). In contrast, PRIMERA demonstrated a relatively moderate improvement with a ROUGE-L increase from 18.90 to 20.48, METEOR increase from 25.15 to 26.50, and PICO-F1 increase from 43.22 to 49.47. However, there was a slight decrease in the CHRF from 39.25 to 37.84. Overall, the fine-tuned LLMs improved the ROUGE-L score by an absolute of 9.89\% (95\% confidence interval of improvement: 8.94-10.81), the METEOR score by 13.21 (95\% confidence interval of improvement: 12.05-14.37), and the CHRF score by 15.82 (95\% confidence interval of improvement: 13.89-16.44). These recently released models all outperformed the fine-tuned variant of BART, the previous SoTA. The fine-tuned BART achieved 17.74, 27.49, and 40.54 in ROUGE-L, METEOR, and CHRF, respectively. We compared the fine-tuned models with GPT-3.5-turbo, one of the most widely known and cutting-edge closed-source LLMs. Zero-shot GPT-3.5-turbo achieved 23.15 in ROUGE-L, 28.83 in METEOR, and 39.74 in CHRF scores. The performance gaps between the open-source models and GPT-3.5-turbo were reduced after fine-tuning. The fine-tuned LongT5 achieved similar results as GPT-3.5-turbo (Supplement Table \ref{tab:automatic}). We also conducted a pilot study using GPT-4. The summaries generated by GPT-3.5-turbo and GPT-4 are not significantly different; as such, we only use GPT-3.5-turbo for comparison.

Furthermore, while few-shot learning helped reduce the performance gap, LLMs fine-tuned with the full training data still demonstrated better performance. We constructed two sets of few-shot learning baselines, one based on few-shot prompting and the other based on few-shot fine-tuning (see details in Methods). Under 1-, 2-, and 5-shot prompting, Mixtral-8x7B achieved 24.87/24.53/24.99 in Rouge-L, 27.82/25.78/27.59 in METEOR, 37.63/35.61/37.42 in CHRF respectively. After fine-tuning using 100 randomly selected samples, PRIMERA achieved 19.11 in ROUGE-L, 24.98 in METEOR, and 39.01 in CHRF; LongT5 was moderately improved with few-shot fine-tuning, resulting in 15.06 in ROUGE-L, 16.17 in METEOR, and 24.81 in CHRF.

We also calculated Pearson correlation coefficients among different automatic metrics using the entire test set (Supplementary Figure~\ref{fig:pearson}). Most metrics are strongly positively correlated with each other; with a few exceptions, most coefficients are larger than 0.7. The only exceptional metric is the PICO coverage, which is only weakly positively correlated to the others, with coefficients ranging between 0.33 and 0.46. As discussed above, zero-shot models tend to verbatim replicate the input, which is already abundant in PICO concepts, since we selected only the objective and main results section of the review abstracts.

\subsection{Comparison between zero-shot LongT5-xl and fine-tuned LongT5-base}\label{comparison-between-zero-shot-longt5-xl-and-fine-tuned-longt5-base}

\begin{figure}
    \centering
    \includegraphics[width=\linewidth]{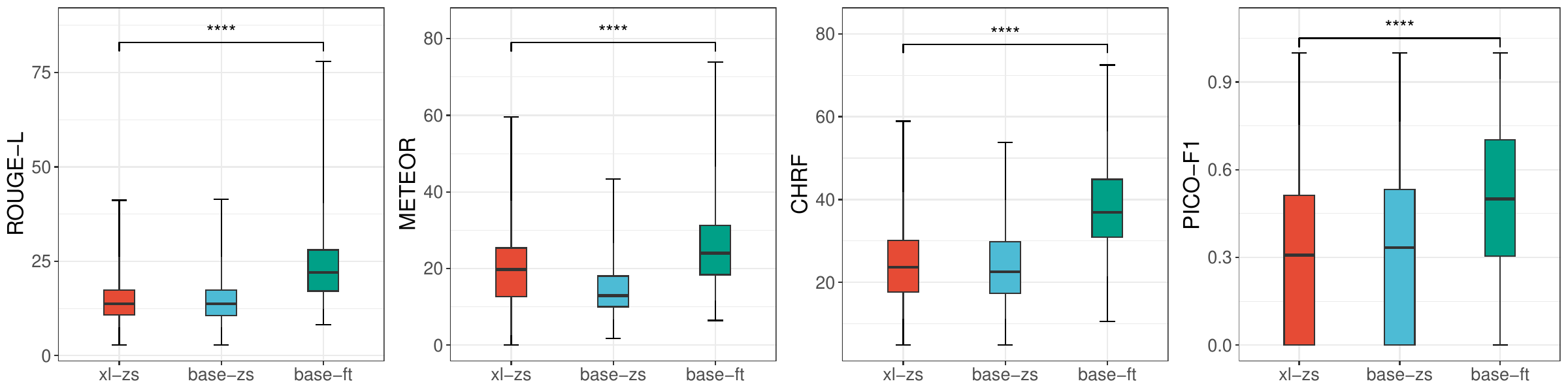}
    \caption{Comparison between zero-shot LongT5-xl and fine-tuned LongT5-base. FT - fine-tuning; ZS - zero-shot learning; * - $p<0.05$; ** - $p<0.01$; *** - $p<0.001$; **** - $p<0.0001$; ns - Not significant.}
    \label{fig:comparison}
\end{figure}

Next, we investigated whether fine-tuned smaller models have the capability to outperform zero-shot larger models. To this end, we fine-tuned the LongT5-base model, which has 10\% fewer parameters than LongT5-xl. Figure~\ref{fig:comparison} shows that fine-tuned LongT5-base outperforms zero-shot LongT5-xl. We also found that this observed trend holds across different LLM architectures. For instance, the performance of fine-tuned PRIMERA and LongT5 exceeded that of zero-shot Llama-2 (Supplementary Table~\ref{tab:automatic}), even though the latter model comprises at least 20 times more parameters.

\subsection{Qualitative evaluation}\label{qualitative-evaluation}

Finally, we conducted a comprehensive human evaluation and a GPT-4 simulated\cite{OpenAI2023-cn} evaluation of machine-generated summaries. Our baseline, zero-shot Llama-2, is one of the latest and largest open-sourced LLMs. In both evaluations, we requested clinical experts or GPT-4 to select the better summary from a pair of candidates - one generated using the baseline and the other generated using our fine-tuned models. The win-rate is the ratio of machine-generated summaries evaluated as better than the baseline. The baseline win-rate is 50\%, given that (1) zero-shot Llama-2 is being compared to itself, (2) all pairs of summaries to be compared are identical, and (3) ties are broken randomly.\cite{Ouyang2022-ww}

\begin{figure}
    \centering
    \includegraphics[width=.7\linewidth]{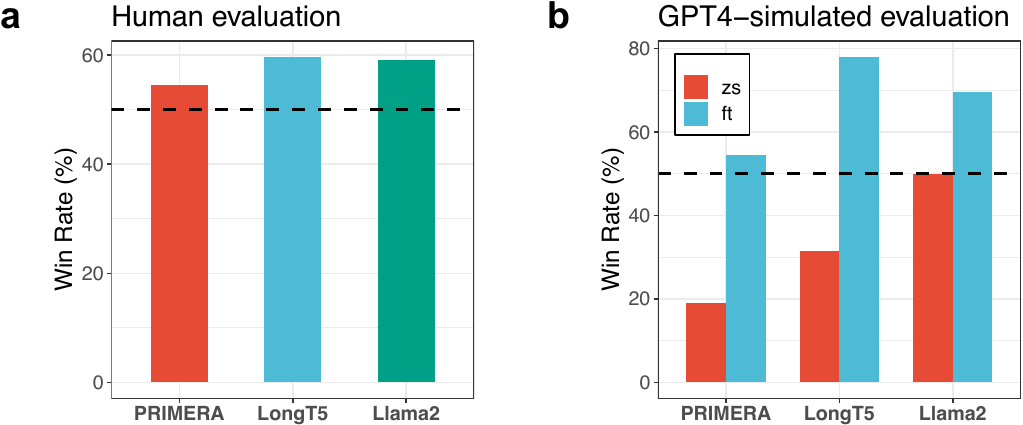}
    \caption{Human and GPT4-simulated evaluation of LLM-generated summaries. \textbf{a,} Performance of different summarization systems in human evaluations using win-rates against zero-shot Llama-2 (Llama-2-zs). The dotted line represents the default 50\% win-rate of the Llama-2-zs. \textbf{b,} Performance of different summarization systems in GPT4-simulated evaluation using win-rate. The dotted line represents the default win-rate of Llama-2-zs. zs - zero-shot learning; ft - fine-tuning.}
    \label{fig:human}
\end{figure}

According to the human evaluation, fine-tuned Llama-2 was preferred to zero-shot, with the win-rate increased from 50\% to 59.20\% (Figure~\ref{fig:human}a, Supplementary Table~\ref{tab:human}). Fine-tuned PRIMERA and LongT5 models also achieved 54.47\% and 59.68\% win-rate against the zero-shot Llama-2.

\begin{figure}
    \centering
    \includegraphics[width=\linewidth]{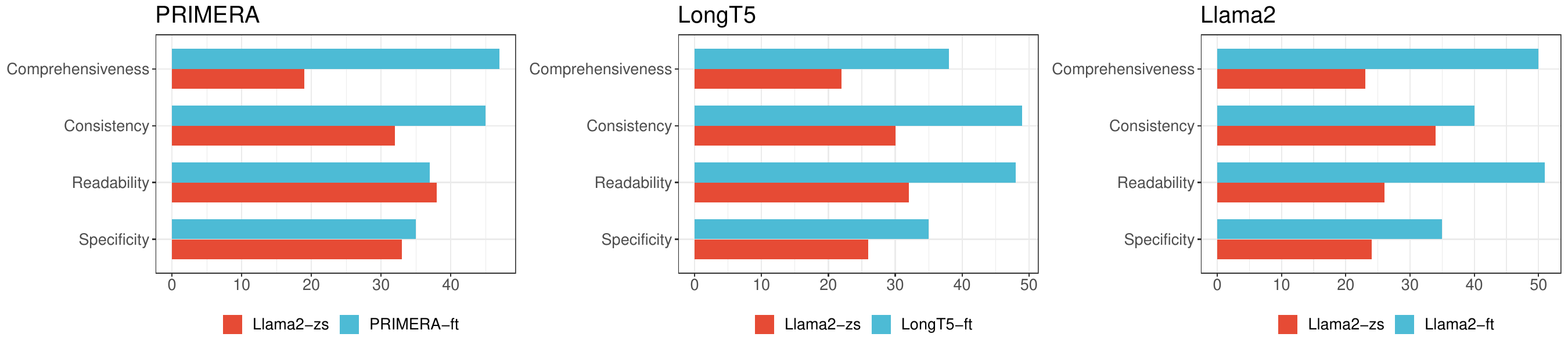}
    \caption{The number of summaries where zero-shot LLama-2 generated better summaries (left/red), in contrast to the cases where the fine-tuned models generated better summaries (right/blue). As compared to zero-shot LLama-2, fine-tuned models produced more comprehensive, readable, consistent, and specific summaries in general. Despite PRIMERA and LongT5 having much smaller model architectures, they significantly outperformed zero-shot LLama-2 after fine-tuning. LLama-2 was also improved in all aspects via fine-tuning.}
    \label{fig:number}
\end{figure}

We further asked the evaluators to share the rationale behind their preference for the chosen summaries. The fine-tuned models were compared with zero-shot Llama-2 on multiple dimensions that have been established as desired properties of summaries.\cite{Tang2023-oa, Fabbri2021-gm} Figure~\ref{fig:number} shows the number of cases where zero-shot LLama-2 generated better summaries (left/red), in contrast to the cases where the fine-tuned models generated better summaries (right/blue). With few exceptions, LLMs were improved in all aspects after fine-tuning (Supplementary Table~\ref{tab:number}). By manually comparing the summaries generated by zero-shot and fine-tuned models, we found that zero-shot models tend to present a detailed background of the summarized studies but do not provide any findings or conclusions, i.e., they present a high resemblance of leading sentences in the paragraphs. Recall that word embeddings are combined with positional embeddings to represent each token in a document in transformer architecture. In general summarization tasks, the key information is typically presented in leading or concluding sentences. The ordering of key information and other redundant information can impact the positional embeddings during pre-training. The zero-shot open-source summarization models mostly extract the leading sentences instead of key information. This indicates that positional embeddings significantly impacted the summary more than word embeddings in zero-shot open-source models. On the other hand, fine-tuned models align more closely with ground-truth summaries, which can provide supportive evidence or identify the lack of sufficient evidence for intervention outcomes.

The GPT-4 simulated evaluation also indicates a significant improvement in all models after fine-tuning (Figure~\ref{fig:human}b). In addition, 257 out of 378 simulated evaluation results concord with the judgment of human experts (68\% accuracy).

We also evaluated all models on two distinct test sets of review articles (Supplementary Table~\ref{tab:simulated}). One group, denoted as ``after cutoff,'' was published after the latest knowledge cutoff date of every model, i.e., no articles in this group were used in pre-training the foundation models. The other group, denoted as ``before cutoff'', was published before the knowledge cutoff date; the articles in this group may be used for pre-training purposes. On both the ``after cutoff'' and the ``before cutoff'' test sets, fine-tuned models demonstrated improved performances. Recap that all articles used for fine-tuning were published ``before cutoff''. This demonstrates the generalizability of fine-tuned models ``before cutoff'' data on ``after cutoff'' test data.

\section{Discussion}\label{discussion}

In this study, we focused on comparing open-source and closed-source LLMs in medical evidence summarization. While closed-source models, exemplified by GPT families and others, demonstrated superior performance as compared to open-source alternatives, the risks associated with using closed-source models are not negligible, which include lack of transparency, reliance on external dependency as a single-point-of-failure, potentially high-cost in migration to other vendors. Furthermore, it's still unclear whether patients would consent to have their information utilized in LLMs, especially when they were unaware of and unable to understand what the LLMs will be used. However, such disadvantages were not paid enough attention. To mitigate the above risks, we investigated recently released open-source LLMs for medical evidence summarization. To our best knowledge, open-source models still fall behind in natural language understanding, as compared to closed-source ones. Based on the observation that fine-tuning can enhance, we validate principles of model optimization within evidence summarization and quantitatively measure the performance boost via fine-tuning.

To facilitate future studies in this direction, we first introduced MedReview, a collection of meta-analysis and summary pairs to assist in fine-tuning LLMs for medical evidence summarization. We further showed that open-source LLMs, including LongT5, PRIMERA, and Llama-2, demonstrated improved summarization performance after fine-tuning with MedReview, even close to GPT-3.5-turbo. This observation also confirms that these generalist models perform well across diverse tasks but need more in-depth domain-specific knowledge, e.g., in biomedicine. While few-shot learning has been effective in other NLP tasks, the applicability is still limited by the context window within lengthy document summarization. Fine-tuning is a robust approach to bridge this performance gap between open-source models and closed-source alternatives, while maintaining the advantages of transparency, easy customization, maintenance, and migration. However, the manual evaluation results show that fine-tuning does not guarantee truthful and accurate summaries. This issue was also reported in the previous evaluation of zero-shot GPT models.\cite{Tang2023-oa} This highlights that trustworthy summarization of medical evidence remains challenging and unresolved.

In addition, the usage of large models is limited by the high demand for computing resources. Our experiments indicated that smaller models, when fine-tuned, can outperform zero-shot larger models on specific tasks. This is observed in two experiments. First, the fine-tuned LongT5-base performs better than the zero-shot LongT5-xl (Figure~\ref{fig:comparison}). Second, PRIMERA and LongT5-xl, despite being smaller in size than Llama-2 (70B), demonstrated enhanced performance than the latter one upon fine-tuning (Figure~\ref{fig:number}, Supplement Table~\ref{tab:automatic}). In the cases of limited computing resources, PRIMERA demonstrated robust performance in medical evidence summarization in the few-shot fine-tuning setup, which confirms the findings of the original reports of the PRIMERA paper. However, few-shot fine-tuned models underperformed as compared to those fully fine-tuned on the entire training data.

The evaluation of LLM-generated summaries presents another challenge in this task. Automatic metrics frequently used for text summarization can, at best, measure the similarity between word distributions of references and LLM-generated summaries. These metrics do not correlate strongly with properties of desired summaries, such as factual comprehensiveness and consistency. As such, human evaluation, especially those from clinical experts, is still critical but not scalable. Therefore, we used GPT-4 as a simulated evaluator and found that 68\% of GPT4-simulated evaluation results aligned with human judgments. These findings suggest that GPT-4 holds the promise to not only facilitate summary evaluation, but also to provide feedback to align the summarization model with simulated expert feedback. The simulated evaluation results further confirmed that fine-tuned smaller models can surpass larger zero-shot models (Figure~\ref{fig:human}b). This observation is consistent with the improvement discussed in the automatic evaluation.

Summarization systems have diverse applications, particularly benefiting researchers and healthcare professionals. For researchers engaged in systematic reviews, these systems can analyze clinical trial reports efficiently, pinpointing relevant studies and distilling essential findings without requiring exhaustive document review. This significantly speeds up the research process. Healthcare professionals and policymakers, on the other hand, can utilize these systems to efficiently grasp the latest clinical trials' outcomes and implications, which is vital for informed decision-making concerning patient care, treatment guidelines, and healthcare policies, especially under the urgent pandemic conditions. Furthermore, by condensing the latest clinical trial information, summarization systems provide concise, current data to clinical decision support systems, assisting healthcare providers in making evidence-based treatment decisions customized to their patients' specific needs.

This study has certain limitations. Due to the restricted time and resources, we cannot extensively explore fine-tuning models such as Claude, GPT-3.5-turbo, and GPT-4. Instead, we mainly focused on open-sourced LLMs. This restriction, however, does not impact the applicability of our study in the clinical domain. We believe that open-sourced LLMs foster wider collaboration and transparency, thus forwarding research progress as other researchers can enhance, modify, and refine these models. Plus, by advocating for the democratization of technology, these LLMs encourage a more extensive use and a potentially higher rate of innovation. Another limitation is that the LLMs were not fine-tuned to summarize clinical trial publications but the manually curated ``main results'' of review abstracts. This study design was initially aimed at simplifying and expediting the development and testing of our summarization algorithms. A future direction is to deploy LLMs to directly synthesize information from clinical trials.

To summarize, our findings underscore the utility of fine-tuning as a robust technique for bridging the performance gap between open-source LLMs and closed-source ones, reinforcing its applicability across a spectrum of model architectures and sizes, and setting the stage for more nuanced investigations into the efficiency and effectiveness of model optimization strategies. However, additional research is warranted to fully uncover the potential of LLM in this context.

\section{Methods}\label{methods}

\subsection{Data Collection}

We collected 8,161 abstracts of systematic reviews from the Cochrane Library.\cite{Cochrane-wg} Unlike abstracts of the biomedical literature, which are highly condensed summaries, systematic review abstracts provide a structured overview that enables readers to quickly determine the validity and applicability of the review. These abstracts typically follow a common structure, detailing preferred reporting items.\cite{Page2021-nl} The Cochrane review abstracts present \emph{background}, \emph{objectives}, \emph{search methods}, \emph{selection criteria}, \emph{data collection and analysis}, \emph{main results}, and \emph{authors' conclusions.} Within such a self-contained structure, the authors' conclusion presents a narrative summary of the most salient details of the included clinical studies.\cite{Tang2023-oa} This section is one of the first to consult when healthcare providers seek answers to clinical questions. Given the meta-analysis results as input, we aim to automatically reproduce this narrative summary. The collected reviews cover a wide range of topics, including but not limited to neurology, gastroenterology, rheumatology, nephrology, and radiology. The publication dates of the reviews range from April 1996 to June 2023.

We split the dataset into distinct training (91.56\%), validation (4.83\%), and test (3.61\%) sets, ensuring that all of the samples appear in one set (Supplementary Table~\ref{tab:dataset}). All LLMs were prepared with a large extent of public textual data collected until a certain moment, known as the cutoff. To maintain a legitimate comparison between LLMs, we put all articles published after September 2022 as test data (because most LLMs studied in this study used the data up to September 2022). Articles published prior to this are primarily used for training and validation. The division of training and validation was stratified according to the time of publication.

\textbf{Few-shot Baselines}

Few-shot learning has been proven an effective and sample-efficient strategy for optimizing task-specific LLMs. We construct two few-shot learning baselines, one based on prompting and the other on fine-tuning.

For few-shot prompting, we use Mixtral-7x8B for the few-shot prompting foundation model. Due to the limit on token numbers, other open-source models cannot fit demonstrations of long document summarization in the context windows. We randomly selected 1, 2, and 5 samples from the training set as demonstrations. For few-shot fine-tuning, we followed the findings of the PRIMERA report that LLMs can be reasonably adapted to domain-specific tasks with a limited number of labeled samples. We used the same setup as the few-shot experiments in the PRIMERA, where we randomly selected 100 samples from the training set and fine-tuned LLMs.

\subsection{Fine-tuning LLMs}\label{fine-tuning-llms}

We investigated several LLM architectures that have recently surfaced for tasks related to summarization tasks or as foundation models. In this study, we only consider models that satisfy the following two conditions. First, models need to be publicly accessible and open-sourced to ensure the transparency and accountability of the models. Second, context windows need to be long enough to digest input without requiring condensation or truncation. Bearing all these factors in mind, we included PRIMERA, LongT5, and Llama-2 in our studies. PRIMERA deploys a pretraining strategy named Entity Pyramid to select and aggregate salient information focusing on document summarization.\cite{Xiao2021-et} LongT5 is an extension of T5 architecture that adopts a summarization pretraining strategy to scale up the input length.\cite{Guo2021-qv} Llama-2 is one of the recently released open-source, scalable foundation models with 7B, 13B, and 70B parameters.\cite{Touvron2023-we} Since Mixtral-8x7B demonstrated similar benchmark performance as Llama-2, we did not fine-tune Mixtral-8x7B.\cite{Jiang2024-jw} Instead, we report the few-shot prompting performance of Mixtral-8x7B.

Following a previous work,\cite{Tang2023-oa} we selected the objective and main results sections of a review abstract as input and used the authors' conclusion as a reference to fine-tune models. We applied the LoRA method, which keeps the original parameters frozen and adjusts only a relatively small number of extra parameters via matrix decomposition.\cite{Hu2021-ey} The exact number of parameters depends on the rank hyperparameter in the LoRA method. We refer the readers to the original LoRA paper for technical details.\cite{Hu2021-ey}

Our implementation uses the following libraries, transformers,\cite{Wolf2020-bs} torch,\cite{Paszke2017-jn} and PEFT.\cite{Mangrulkar2022-sv} Most fine-tuning jobs were completed on AWS and our local lab servers. Llama-2 models were fine-tuned on SageMaker platform. All models were fine-tuned for 1 epoch, within which the validation loss had already stopped decreasing. We set the learning rates to 3e-5,\cite{Xiao2021-et} 1e-4,\cite{Touvron2023-we} and 1e-3\cite{Guo2021-qv} for PRIMERA, LLama-2, and LongT5 models, respectively. We set the rank hyperparameter of LoRA to 8.\cite{Hu2021-ey}

\subsection{Evaluation Metrics}\label{evaluation-metrics}

We first use the natural language generation (NLG) metrics to evaluate the quality of the generated summary. These metrics include ROUGE-L (Recall-Oriented Understudy for Gisting Evaluation) and METEOR (Metric for Evaluation of Translation with Explicit Ordering) scores. We also include the CHRF (CHaRacter-level F-score), which was reported to correlate highly with the readability of generated text.\cite{Fabbri2021-gm} Their values range from 0.0 to 100.0, with a score of 100.0 indicating that the generated summaries are identical to the reference summary. The model performance on our test is approximately normally distributed and the performance of each model is independent of others. As such, we calculated the p-value using a paired t-test to determine the statistical significance of the difference between the two models.

\subsubsection{PICO metrics}\label{pico-metrics}

NLG metrics are known to be inadequate for evaluating factual completeness and consistency.\cite{Fabbri2021-gm} We therefore propose to use a PICO (Participants, Interventions, Comparison, and Outcomes) extraction system to evaluate the accuracy of the generated summaries. More specifically, we fine-tuned a BERT-based\cite{Gu2021-ni} model to extract PICO concepts and then score a generated summary by comparing the values of these PICO elements obtained from the reference. We consider a PICO element to be a true positive, if it satisfies two conditions: (1) the text in the reference overlaps with the text in the generation, and (2) the two entity types should have the same PICO category. The micro averages for precision, recall, and F1 scores are all computed over the PICO components.

\subsubsection{Human evaluation}\label{human-evaluation}

We conducted a review of the summary quality via human evaluation. The quality is measured from four aspects: consistency, comprehensiveness, specificity, and readability, which were established as essential factors for measuring machine summary quality. Consistency indicates whether the summary contradicts the input source. Comprehensiveness measures coverage of key information of input. Specificity measures the preciseness and conciseness of the summary. Readability indicates a machine summary is fluent and free of grammatical errors that hinder understanding.

To evaluate the machine summaries, we invited 7 clinical experts, each specializing in one or two of the following specialties, including Gastroenterology, General Surgery, Internal Medicine, Nephrology, Neurology, Radiology, and Rheumatology. All experts have obtained MD training and currently provide direct patient care. Following a recent LLM study,\cite{Ouyang2022-ww} we request the experts to compare the quality of different machine summaries (interface shown in Supplementary Figure~\ref{fig:user interface}). Specifically, according to their domain knowledge, each expert was assigned review abstracts along with three summaries: (1) the Authors' conclusion section, (2) zero-shot Llama-2, and (3) one of the fine-tuned models. From the zero-shot baseline and the fine-tuned models, the experts will select which one generates a better summary. To reduce the potential order-related bias, the order of summaries generated by zero-shot Llama-2 and fine-tuned summaries was randomized. We further asked the experts to choose the reasons for their choices.

\subsubsection{GPT-4 evaluation}\label{gpt-4-evaluation}

Like many other annotation scenarios, collecting experts' feedback is not scalable with respect to the samples to be labeled. In addition to our manual review, we explored the use of GPT-4\cite{OpenAI2023-cn} as a simulated expert to answer the same questions as those assigned to human experts. Instead of selecting a sample of test data as in the manual review, we used the model summaries for all test articles for GPT-4 evaluation. We also analyzed the percentage of questions that human judgments agree with the GPT-4 evaluation.

\section*{Data availability}

The data can be accessed at \url{https://github.com/ebmlab/MedReview}.

\section*{Code Availability}

The code can be accessed at \url{https://github.com/ebmlab/MedReview}.

\section*{Acknowledgements}\label{acknowledgements}

This project was sponsored by the National Library of Medicine grant R01LM009886, R01LM014344, National Human Genome Research Institute grant R01HG012655, and the National Center for Advancing Clinical and Translational Science awards UL1TR001873 and UL1TR002384. Q.J. and Z.L. are supported by the NIH Intramural Research Program, National Library of Medicine. We also want to express our gratitude to Amazon Web Services (AWS) for providing the computational resources used in our research. The funder had no role in the design and conduct of the study; collection, management, analysis, and interpretation of the data; preparation, review, or approval of the manuscript; and decision to submit the manuscript for publication.

\section*{Author Contributions:} 

Study concepts/study design, \textbf{G.Z., C.W., Y.P.}; manuscript drafting or manuscript revision for important intellectual content, \textbf{G.Z., Q.J., Y.Z., S.W., B.R.I., Y.L., E.P., J.G.N., M.E.S., A.S., T.C., Z.L., C.W., Y.P.}; approval of final version of the submitted manuscript, \textbf{G.Z., Q.J., Y.Z., S.W., B.R.I., Y.L., E.P., J.G.N., M.E.S., A.S., T.C., Z.L., C.W., Y.P.}; agrees to ensure any questions related to the work are appropriately resolved, \textbf{G.Z., Q.J., Y.Z., S.W., B.R.I., Y.L., E.P., J.G.N., M.E.S., A.S., T.C., Z.L., C.W., Y.P.}; literature research, \textbf{G.Z., Y.P.}; experimental studies, \textbf{G.Z., Q.J., Y.Z., S.W.}; human evaluation, \textbf{Q.J., B.R.I., Y.L., E.P., J.G.N., M.E.S., A.S.}; data interpretation and statistical analysis, \textbf{G.Z., Y.P.}; and manuscript editing, \textbf{G.Z., Q.J., Y.Z., S.W., B.R.I., Y.L., E.P., J.G.N., M.E.S., A.S., T.C., Z.L., C.W., Y.P.}.

\section*{Competing Interests}\label{competing-interests}

The authors declare no competing interests.

\section*{Ethics declarations}\label{ethics-declarations}

Not applicable.

\bibliographystyle{medline}
\bibliography{ref}

\begin{thebibliography}{10}

\bibitem{Peng2023-zh}
Peng Y, Rousseau JF, Shortliffe EH, Weng C.
\newblock {AI-generated} text may have a role in evidence-based medicine.
\newblock Nat Med. 2023 Jul;29(7):1593--1594.
\newblock doi: 10.1038/s41591-023-02366-9. PMID: 37221382. PMCID: 6259661.

\bibitem{Concato2000-np}
Concato J, Shah N, Horwitz RI.
\newblock Randomized, controlled trials, observational studies, and the hierarchy of research designs.
\newblock N Engl J Med. 2000 Jun;342(25):1887--1892.
\newblock doi: 10.1056/NEJM200006223422507. PMID: 10861325. PMCID: PMC1557642.

\bibitem{Borah2017-ah}
Borah R, Brown AW, Capers PL, Kaiser KA.
\newblock Analysis of the time and workers needed to conduct systematic reviews of medical interventions using data from the {PROSPERO} registry.
\newblock BMJ Open. 2017 Feb;7(2):e012545.
\newblock doi: 10.1136/bmjopen-2016-012545. PMID: 28242767. PMCID: PMC5337708.

\bibitem{NLM2024-qw}
NLM. {ClinicalTrial.gov} ({NLM)}, Accessed on Nov 18, 2023.
\newblock Accessed: 2023-11-18.

\bibitem{Page2021-nl}
Page MJ, McKenzie JE, Bossuyt PM, Boutron I, Hoffmann TC, Mulrow CD, Shamseer L, Tetzlaff JM, Akl EA, Brennan SE, Chou R, Glanville J, Grimshaw JM, Hr{\'o}bjartsson A, Lalu MM, Li T, Loder EW, Mayo-Wilson E, McDonald S, McGuinness LA, Stewart LA, Thomas J, Tricco AC, Welch VA, Whiting P, Moher D.
\newblock The {PRISMA} 2020 statement: an updated guideline for reporting systematic reviews.
\newblock Rev Esp Cardiol. 2021 Sep;74(9):790--799.
\newblock doi: 10.1016/j.rec.2021.07.010. PMID: 34446261.

\bibitem{Wallace2021-zh}
Wallace BC, Saha S, Soboczenski F, Marshall IJ.
\newblock Generating (Factual?) Narrative Summaries of {RCTs}: Experiments with Neural {Multi-Document} Summarization.
\newblock AMIA Jt Summits Transl Sci Proc. 2021 May;2021:605--614.
\newblock PMID: 34457176. PMCID: PMC8378607.

\bibitem{Tang2023-oa}
Tang L, Sun Z, Idnay B, Nestor JG, Soroush A, Elias PA, Xu Z, Ding Y, Durrett G, Rousseau JF, Weng C, Peng Y.
\newblock Evaluating large language models on medical evidence summarization.
\newblock NPJ Digit Med. 2023 Aug;6(1):158.
\newblock doi: 10.1038/s41746-023-00896-7. PMID: 37620423. PMCID: PMC10449915.

\bibitem{Barzilay2002-nv}
Barzilay R, Elhadad N.
\newblock Inferring Strategies for Sentence Ordering in Multidocument News Summarization.
\newblock jair. 2002 Aug;17:35--55.
\newblock doi: 10.1613/jair.991.

\bibitem{Pivovarov2015-km}
Pivovarov R, Elhadad N.
\newblock Automated methods for the summarization of electronic health records.
\newblock J Am Med Inform Assoc. 2015 Sep;22(5):938--947.
\newblock doi: 10.1093/jamia/ocv032. PMID: 25882031. PMCID: PMC4986665.

\bibitem{Zweigenbaum2007-jj}
Zweigenbaum P, Demner-Fushman D, Yu H, Cohen KB.
\newblock Frontiers of biomedical text mining: current progress.
\newblock Brief Bioinform. 2007 Sep;8(5):358--375.
\newblock doi: 10.1093/bib/bbm045. PMID: 17977867. PMCID: PMC2516302.

\bibitem{Li2010-rc}
Li F, Han C, Huang M, Zhu X, Xia YJ, Zhang S, Yu H.
\newblock {Structure-Aware} Review Mining and Summarization.
\newblock In: Huang CR, Jurafsky D, editors. Proceedings of the 23rd International Conference on Computational Linguistics (Coling 2010). Beijing, China: Coling 2010 Organizing Committee; 2010. p. 653--661.

\bibitem{Demner-Fushman2007-tk}
Demner-Fushman D, Lin JJ.
\newblock Answering clinical questions with knowledge-based and statistical techniques.
\newblock CL. 2007 Mar;33:63--103.
\newblock doi: 10.1162/coli.2007.33.1.63.

\bibitem{Gu2021-ni}
Gu Y, Tinn R, Cheng H, Lucas M, Usuyama N, Liu X, Naumann T, Gao J, Poon H.
\newblock {Domain-Specific} Language Model Pretraining for Biomedical Natural Language Processing.
\newblock ACM Trans Comput Healthcare. 2021 Oct;3(1):1--23.
\newblock doi: 10.1145/3458754.

\bibitem{Guo2021-qv}
Guo M, Ainslie J, Uthus D, Ontanon S, Ni J, Sung YH, Yang Y.
\newblock {LongT5}: Efficient text-to-text transformer for long sequences.
\newblock In: Findings of the Association for Computational Linguistics: NAACL 2022. Stroudsburg, PA, USA: Association for Computational Linguistics; 2022. p. 724–736.
\newblock doi: 10.18653/v1/2022.findings-naacl.55.

\bibitem{Xiao2021-et}
Xiao W, Beltagy I, Carenini G, Cohan A.
\newblock {PRIMERA}: Pyramid-based Masked Sentence Pre-training for Multi-document Summarization.
\newblock In: Proceedings of the 60th Annual Meeting of the Association for Computational Linguistics (Volume 1: Long Papers). Stroudsburg, PA, USA: Association for Computational Linguistics; 2022. p. 5245–5263.
\newblock doi: 10.18653/v1/2022.acl-long.360.

\bibitem{Zhang2020-fq}
Zhang J, Zhao Y, Saleh M, Liu P.
\newblock {{PEGASUS}}: Pre-training with Extracted Gap-sentences for Abstractive Summarization.
\newblock In: Iii HD, Singh A, editors. Proceedings of the 37th International Conference on Machine Learning. vol. 119 of Proceedings of Machine Learning Research. PMLR; 2020. p. 11328--11339.

\bibitem{Lewis2019-qk}
Lewis M, Liu Y, Goyal N, Ghazvininejad M, Mohamed A, Levy O, Stoyanov V, Zettlemoyer L.
\newblock {BART}: Denoising Sequence-to-Sequence Pre-training for Natural Language Generation, Translation, and Comprehension.
\newblock In: Proceedings of the 58th Annual Meeting of the Association for Computational Linguistics. Online: Association for Computational Linguistics; 2020. p. 7871--7880.
\newblock doi: 10.18653/v1/2020.acl-main.703.

\bibitem{Devlin2018-ct}
Devlin J, Chang MW, Lee K, Toutanova K.
\newblock {BERT}: Pre-training of Deep Bidirectional Transformers for Language Understanding.
\newblock In: Proceedings of the 2019 Conference of the North {A}merican Chapter of the Association for Computational Linguistics: Human Language Technologies, Volume 1 (Long and Short Papers). Minneapolis, Minnesota: Association for Computational Linguistics; 2019. p. 4171--4186.
\newblock doi: 10.18653/v1/N19-1423.

\bibitem{Mrabet2020-xj}
Mrabet Y, Demner-Fushman D.
\newblock {{HOLMS}}: Alternative Summary Evaluation with Large Language Models.
\newblock In: Scott D, Bel N, Zong C, editors. Proceedings of the 28th International Conference on Computational Linguistics. Barcelona, Spain (Online): International Committee on Computational Linguistics; 2020. p. 5679--5688.
\newblock doi: 10.18653/v1/2020.coling-main.498.

\bibitem{Singhal2023-jc}
Singhal K, Azizi S, Tu T, Mahdavi SS, Wei J, Chung HW, Scales N, Tanwani A, Cole-Lewis H, Pfohl S, Payne P, Seneviratne M, Gamble P, Kelly C, Babiker A, Sch{\"a}rli N, Chowdhery A, Mansfield P, Demner-Fushman D, Ag{\"u}era Y~Arcas B, Webster D, Corrado GS, Matias Y, Chou K, Gottweis J, Tomasev N, Liu Y, Rajkomar A, Barral J, Semturs C, Karthikesalingam A, Natarajan V.
\newblock Large language models encode clinical knowledge.
\newblock Nature. 2023 Aug;620(7972):172--180.
\newblock doi: 10.1038/s41586-023-06291-2. PMID: 37438534. PMCID: PMC10396962.

\bibitem{Zack2023-jd}
Zack T, Lehman E, Suzgun M, Rodriguez JA, Celi LA, Gichoya J, Jurafsky D, Szolovits P, Bates DW, Abdulnour REE, Butte AJ, Alsentzer E.
\newblock Coding inequity: Assessing {GPT-4's} potential for perpetuating racial and gender biases in healthcare.
\newblock bioRxiv. 2023 Jul.
\newblock doi: 10.1101/2023.07.13.23292577.

\bibitem{Jin2023-pk}
Jin Q, Yang Y, Chen Q, Lu Z.
\newblock {GeneGPT}: Augmenting Large Language Models with Domain Tools for Improved Access to Biomedical Information.
\newblock ArXiv. 2023 May.
\newblock PMID: 37131884. PMCID: PMC10153281.

\bibitem{Jiang2024-jw}
Jiang AQ, Sablayrolles A, Roux A, Mensch A, Savary B, Bamford C, Chaplot DS, Casas Ddl, Hanna EB, Bressand F, Lengyel G, Bour G, Lample G, Lavaud LR, Saulnier L, Lachaux MA, Stock P, Subramanian S, Yang S, Antoniak S, Scao TL, Gervet T, Lavril T, Wang T, Lacroix T, Sayed WE.
\newblock Mixtral of Experts.
\newblock arXiv. 2024 Jan.

\bibitem{Ouyang2022-ww}
Ouyang L, Wu J, Jiang X, Almeida D, Wainwright C, Mishkin P, Zhang C, Agarwal S, Slama K, Ray A, Schulman J, Hilton J, Kelton F, Miller L, Simens M, Askell A, Welinder P, Christiano PF, Leike J, Lowe R.
\newblock Training language models to follow instructions with human feedback.
\newblock In: Koyejo S, Mohamed S, Agarwal A, Belgrave D, Cho K, Oh A, editors. Advances in Neural Information Processing Systems. vol.~35. Curran Associates, Inc.; 2022. p. 27730--27744.

\bibitem{Touvron2023-we}
Touvron H, Martin L, Stone K, Albert P, Almahairi A, Babaei Y, Bashlykov N, Batra S, Bhargava P, Bhosale S, Bikel D, Blecher L, Ferrer CC, Chen M, Cucurull G, Esiobu D, Fernandes J, Fu J, Fu W, Fuller B, Gao C, Goswami V, Goyal N, Hartshorn A, Hosseini S, Hou R, Inan H, Kardas M, Kerkez V, Khabsa M, Kloumann I, Korenev A, Koura PS, Lachaux MA, Lavril T, Lee J, Liskovich D, Lu Y, Mao Y, Martinet X, Mihaylov T, Mishra P, Molybog I, Nie Y, Poulton A, Reizenstein J, Rungta R, Saladi K, Schelten A, Silva R, Smith EM, Subramanian R, Tan XE, Tang B, Taylor R, Williams A, Kuan JX, Xu P, Yan Z, Zarov I, Zhang Y, Fan A, Kambadur M, Narang S, Rodriguez A, Stojnic R, Edunov S, Scialom T.
\newblock Llama 2: Open Foundation and {Fine-Tuned} Chat Models.
\newblock arXiv. 2023 Jul.

\bibitem{OpenAI2023-cn}
OpenAI R.
\newblock Gpt-4 technical report. arxiv 2303.08774.
\newblock View in Article. 2023.

\bibitem{Nosek2015-yy}
Nosek BA, Alter G, Banks GC, Borsboom D, Bowman SD, Breckler SJ, Buck S, Chambers CD, Chin G, Christensen G, Contestabile M, Dafoe A, Eich E, Freese J, Glennerster R, Goroff D, Green DP, Hesse B, Humphreys M, Ishiyama J, Karlan D, Kraut A, Lupia A, Mabry P, Madon T, Malhotra N, Mayo-Wilson E, McNutt M, Miguel E, Paluck EL, Simonsohn U, Soderberg C, Spellman BA, Turitto J, VandenBos G, Vazire S, Wagenmakers EJ, Wilson R, Yarkoni T.
\newblock Promoting an open research culture.
\newblock Science. 2015;348(6242):1422--1425.
\newblock doi: 10.1126/science.aab2374.

\bibitem{Zhang2024-bu}
Zhang G, Jin Q, Jered~McInerney D, Chen Y, Wang F, Cole CL, Yang Q, Wang Y, Malin BA, Peleg M, Wallace BC, Lu Z, Weng C, Peng Y.
\newblock Leveraging generative {AI} for clinical evidence synthesis needs to ensure trustworthiness.
\newblock J Biomed Inform. 2024 May;153:104640.
\newblock doi: 10.1016/j.jbi.2024.104640. PMID: 38608915.

\bibitem{Gutierrez2022-xu}
Jimenez~Gutierrez B, McNeal N, Washington C, Chen Y, Li L, Sun H, Su Y.
\newblock Thinking about {GPT}-3 in-context learning for biomedical {IE}? Think again.
\newblock In: Findings of the Association for Computational Linguistics: EMNLP 2022. Stroudsburg, PA, USA: Association for Computational Linguistics; 2022. p. 4497–4512.
\newblock doi: 10.18653/v1/2022.findings-emnlp.329.

\bibitem{Tadros2022-ul}
Tadros T, Krishnan GP, Ramyaa R, Bazhenov M.
\newblock Sleep-like unsupervised replay reduces catastrophic forgetting in artificial neural networks.
\newblock Nat Commun. 2022 Dec;13(1):7742.
\newblock doi: 10.1038/s41467-022-34938-7. PMID: 36522325. PMCID: PMC9755223.

\bibitem{Hu2021-ey}
Hu EJ, Shen Y, Wallis P, Allen-Zhu Z, Li Y, Wang S, Wang L, Chen W.
\newblock {LoRA}: {Low-Rank} Adaptation of Large Language Models.
\newblock arXiv. 2021 Jun.

\bibitem{Cochrane-wg}
Cochrane. {The Cochrane Library}.
\newblock Accessed: 2023-9-30.

\bibitem{Fabbri2021-gm}
Fabbri AR, Kry{\'s}ci{\'n}ski W, McCann B, Xiong C, {others}.
\newblock Summeval: Re-evaluating summarization evaluation.
\newblock Transactions of the. 2021.

\bibitem{Wolf2020-bs}
Wolf T, Debut L, Sanh V, Chaumond J, Delangue C, Moi A, Cistac P, Rault T, Louf R, Funtowicz M, Davison J, Shleifer S, von Platen P, Ma C, Jernite Y, Plu J, Xu C, Le~Scao T, Gugger S, Drame M, Lhoest Q, Rush A.
\newblock Transformers: {State-of-the-Art} Natural Language Processing.
\newblock In: Liu Q, Schlangen D, editors. Proceedings of the 2020 Conference on Empirical Methods in Natural Language Processing: System Demonstrations. Online: Association for Computational Linguistics; 2020. p. 38--45.
\newblock doi: 10.18653/v1/2020.emnlp-demos.6.

\bibitem{Paszke2017-jn}
Paszke A, Gross S, Chintala S, Chanan G, Yang E, DeVito Z, Lin Z, Desmaison A, Antiga L, Lerer A.
\newblock Automatic differentiation in {PyTorch}.
\newblock In: Neural Information Processing Systems (NIPS 2017); 2017. p. 1--4.

\bibitem{Mangrulkar2022-sv}
Mangrulkar S, Gugger S, Debut L, Belkada Y, Paul S, Bossan B. {PEFT}: State-of-the-art {Parameter-Efficient} {Fine-Tuning} methods.

\end{thebibliography}

\newpage
\appendix
\setcounter{table}{0}
\setcounter{figure}{0}
\renewcommand\figurename{Supplementary Figure} 
\renewcommand\tablename{Supplementary Table}


\begin{figure}[!htbp]
\centering
\includegraphics[width=\linewidth]{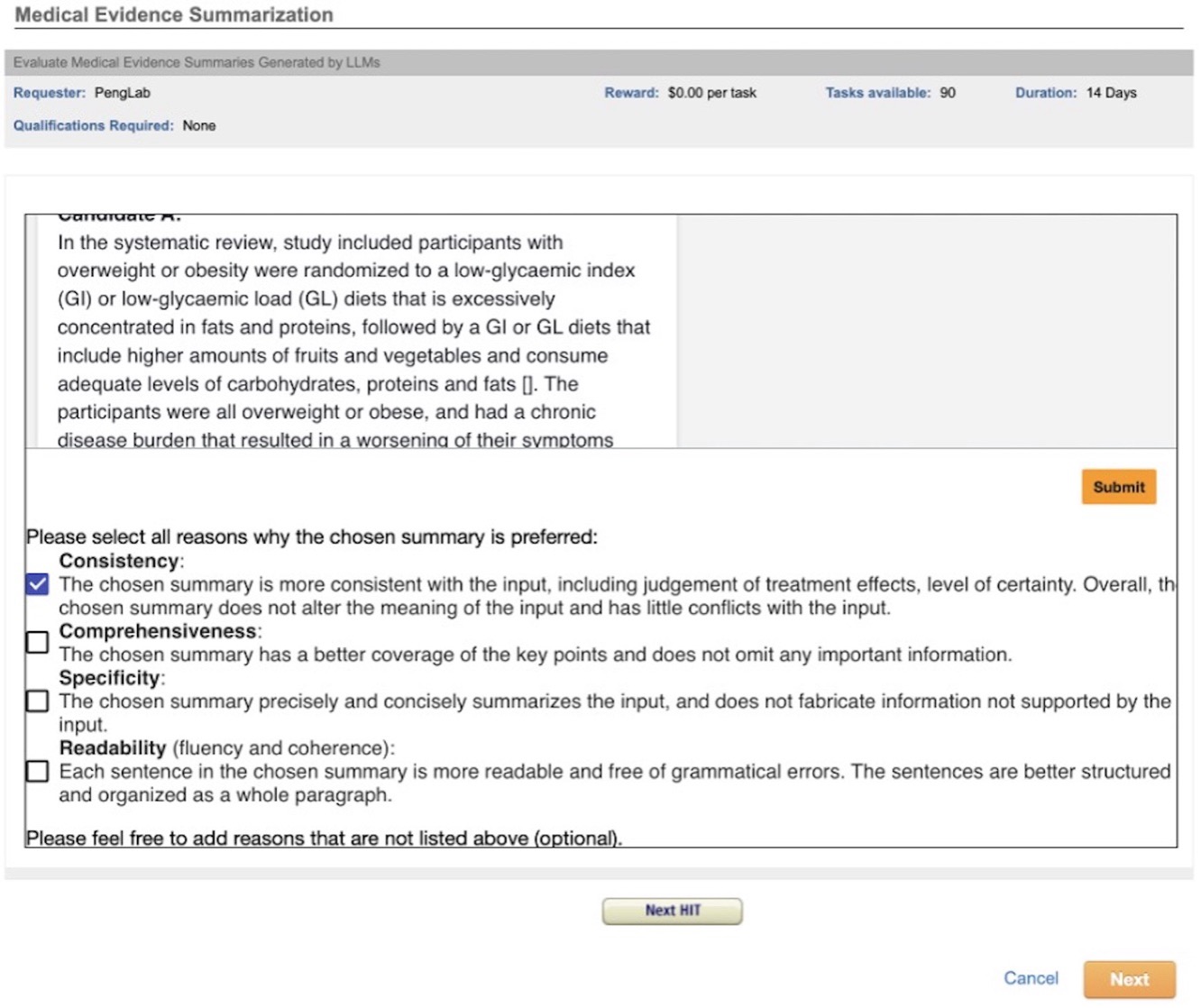}
\caption{User interface for collecting human feedback. The upper box shows an example summary produced by a studied LLM. The lower box displays the multiple choice question about the rationale of the human evaluators’ preference.}
\label{fig:user interface}
\end{figure}

\newpage

\begin{figure}[!htbp]
\centering
\includegraphics[width=.5\linewidth]{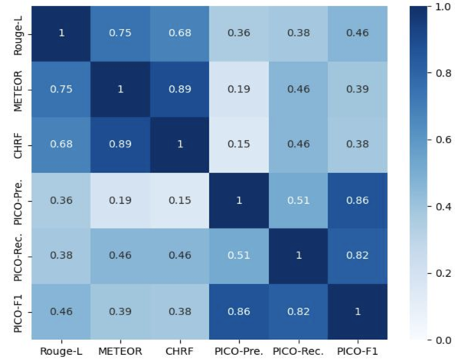}
\caption{Pearson Correlation Coefficients (r) among evaluation metrics. The natural language generation (NLG) metrics have a strongly positive correlation between each other ($r>0.68$). The PICO metrics have a moderate positive correlation with NLG metrics, ($0.15<r<0.46$). Recall that NLG metrics focus on lexical similarity while PICO metrics focus on coverage of key information (PICO elements in the summary).}
\label{fig:pearson}
\end{figure}

\newpage

\begin{table}[!htbp]
\centering
\caption{Dataset used for medical evidence summarization task.}
\label{tab:dataset}
\begin{tabularx}{.4\linewidth}{Xr}
\toprule
\textbf{Dataset} & \textbf{$n$} \\
\midrule
Training & 7,472 \\
Validation & 394 \\
Test & 295 \\
\bottomrule
\end{tabularx}
\end{table}

\newpage

\begin{table}[!htbp]
\centering
\caption{Automatic evaluation scores of LLMs.}
\label{tab:automatic}
\begin{tabular}{lcccccccccc}
\toprule
& \textbf{BART} & \textbf{GPT-3.5} & \multicolumn{2}{c}{\textbf{PRIMERA}} & \multicolumn{2}{c}{\textbf{LongT5-base}} & \multicolumn{2}{c}{\textbf{LongT5-xl}} & \multicolumn{2}{c}{\textbf{Llama-2}} \\
\cmidrule(rl){2-2}\cmidrule(rl){3-3}\cmidrule(rl){4-5}\cmidrule(rl){6-7}\cmidrule(rl){8-9}\cmidrule(rl){10-11}
\textbf{Metrics} & FT & ZS & ZS & FT & ZS & FT & ZS & FT & ZS & FT \\
\midrule
ROUGE-L & 17.74 & 23.15 & 18.90 & 20.48 & 14.67 & 23.66 & 14.72 & \textbf{24.61} & 17.21 & 19.91 \\
METEOR & 27.49 & \textbf{28.83} & 25.15 & 26.50 & 14.70 & 25.93 & 15.06 & 28.27 & 19.69 & 25.02 \\
CHRF & \textbf{40.54} & 39.74 & 39.25 & 37.84 & 22.24 & 36.38 & 22.99 & 38.81 & 30.24 & 36.42 \\
PICO Precision & 42.29 & 48.61 & 34.86 & 49.18 & 52.61 & 53.32 & 49.73 & 53.76 & 50.83 & \textbf{55.28} \\
PICO Recall & 59.63 & \textbf{66.40} & 56.88 & 49.77 & 31.77 & 54.36 & 31.25 & 60.21 & 45.58 & 48.97 \\
PICO F1 & 49.49 & 56.41 & 43.22 & 49.47 & 39.61 & 53.83 & 38.38 & \textbf{56.80} & 48.07 & 51.93 \\
\bottomrule
\end{tabular}
\end{table}

\newpage

\begin{table}[!htbp]
\centering
\caption{Human evaluation of fine-tuned models.}
\label{tab:human}
\resizebox{\linewidth}{!}{
\begin{tabular}{lrrrrrrr}
\toprule
\textbf{Model}
 & \makecell[r]{\textbf{AD/Dementia/}\\\textbf{Neurology}}
 & \makecell[r]{\textbf{Gastro-}\\\textbf{enterology}}
 & \makecell[r]{\textbf{Internal}\\\textbf{Medicine}}
 & \textbf{Nephrology}
 & \makecell[r]{\textbf{Rheuma-}\\\textbf{tology}}
 & \textbf{Surgery}
 & \textbf{Overall}\\
\midrule
PRIMERA & 12/18 & 13/18 & 24/36 & 9/18 & 5/18 & 6/18 & (69/126) 54.76\% \\
LongT5 & 12/18 & 12/18 & 25/36 & 13/18 & 6/18 & 7/18 & (75/124) 59.52\% \\
Llama-2 & 15/18 & 10/18 & 24/36 & 8/18 & 9/18 & 8/18 & (75/126) 58.73\% \\
\bottomrule
\end{tabular}
}
\end{table}

\newpage

\begin{table}[!htbp]
\centering
\caption{Simulated evaluation by GPT-4.}
\label{tab:simulated}
\begin{tabularx}{.5\linewidth}{Xr@{}lr@{}l}
\toprule
\textbf{Model} & \textbf{Before Cutoff} && \textbf{After Cutoff} & \\
\midrule
Llama-2		\\
\hspace{1em}ZS & 50.00&* & 50.00&*\\
\hspace{1em}FT & 67.39& & 69.49\\
PRIMERA		\\
\hspace{1em}ZS & 19.08& & 18.98\\
\hspace{1em}FT & 55.56& & 54.58\\
LongT5		\\
\hspace{1em}ZS & 35.02& & 31.53\\
\hspace{1em}FT & 74.88& & 77.97\\
\bottomrule
\end{tabularx}
\end{table}

\newpage

\begin{table}[!htbp]
\centering
\caption{The number of summaries with better qualities.}
\label{tab:number}
\begin{tabular}{llrrrr}
\toprule
&&\multicolumn{4}{c}{\textbf{\# of summaries with better qualities}}\\
\cmidrule{3-6}
&&\multicolumn{2}{c}{\textbf{Zero-shot Llama-2}} & 
\multicolumn{2}{c}{\textbf{Fine-tuned models}}\\
\cmidrule(rl){3-4}\cmidrule(rl){5-6}
\textbf{Model} & \textbf{Desired Properties} & $n$ & \% & $n$ & \% \\
\midrule
{PRIMERA (ft)} & Comprehensiveness & 19 & (15.08\%) & 47 & (37.30\%) \\
& Consistency & 32 & (25.40\%) & 45 & (35.71\%) \\
& Readability & 38 & (30.16\%) & 37 & (29.37\%) \\
& Specificity & 33 & (26.19\%) & 35 & (27.78\%) \\
\midrule
{LongT5-xl (ft)} & Comprehensiveness & 22 & (17.46\%) & 38 & (30.16\%) \\
& Consistency & 30 & (23.81\%) & 49 & (38.89\%) \\
& Readability & 32 & (25.40\%) & 48 & (38.10\%) \\
& Specificity & 26 & (20.63\%) & 35 & (27.78\%) \\
\midrule
{Llama-2 (ft)} & Comprehensiveness & 23 & (18.25\%) & 50 & (39.68\%) \\
& Consistency & 34 & (26.98\%) & 40 & (31.75\%) \\
& Readability & 36 & (28.57\%) & 52 & (41.27\%) \\
& Specificity & 24 & (19.05\%) & 35 & (27.78\%) \\
\bottomrule
\end{tabular}
\end{table}

\newpage

\begin{table}[!htbp]
\centering
\caption{Summaries generated by zero-shot, fine-tuned LongT5, GPT-3.5-turbo, input article doi: 10.1002/14651858.CD012458.}
\label{tab:example}
\begin{tabularx}{\linewidth}{lX}
\toprule
Model & Summary\\
\midrule
LongT5(zero shot) & Our secondary objective was to assess the effect of single-room isolation or cohorting, or both on neonatal mortality and perceived or documented adverse effects in newborn infants admitted to the NICU.\\
\midrule
LongT5(fine-tuned) & We found no evidence to support or refute the use of single-room isolation or cohorting for preventing transmission of HAIs or colonization with HAI causing pathogens in newborn infants less than six months of age admitted to the neonatal intensive care unit (NICU).\\
\midrule
GPT-3.5-turbo & This systematic review aimed to evaluate the impact of single-room isolation or cohorting, or both, on the transmission of healthcare-associated infections (HAIs) in newborn infants admitted to the neonatal intensive care unit (NICU). However, no published or ongoing trials were found to include in the review. Therefore, no conclusions could be drawn regarding the effectiveness of these isolation measures on preventing HAIs or colonization with HAI-causing pathogens in this population. Further research is needed in this area.\\
\midrule
Human Expert (Reference) & The review found no evidence from randomized trials to either support or refute the use of patient isolation measures (single‐room isolation or cohorting) in neonates with HAIs. Risks secondary to infection control measures need to be balanced against the benefits of decreasing horizontal transmission in the neonatal unit for optimal neonatal outcomes. There is an urgent need to research the effectiveness of patient isolation measures for preventing the transmission of HAIs in neonatal units. Well‐designed trials randomizing clusters of units or hospitals to a type of patient isolation method intervention are warranted.\\
\bottomrule
\end{tabularx}
\end{table}

\end{document}